\documentclass{article}

\usepackage{microtype}
\usepackage{graphicx}
\usepackage{subfigure}
\usepackage{booktabs} 

\usepackage{hyperref}

\usepackage[accepted]{icml2025}

\usepackage{amsmath}
\usepackage{amssymb}
\usepackage{mathtools}
\usepackage{amsthm}

\usepackage[capitalize,noabbrev]{cleveref}

\theoremstyle{plain}

\theoremstyle{definition}

\theoremstyle{remark}

\usepackage[textsize=tiny]{todonotes}

\usepackage{adjustbox}
\usepackage{xspace}
\usepackage{amssymb}
\usepackage{multirow}
\usepackage[acronym]{glossaries}
\usepackage[inline]{enumitem}
\usepackage{graphicx}
\usepackage{subcaption}
\usepackage{caption}
\usepackage{booktabs}
\usepackage{setspace}
\usepackage{tabularx}
\usepackage{wrapfig}
\usepackage{afterpage} 
\usepackage{setspace}
\usepackage{float}     

\usepackage[utf8]{inputenc}
\usepackage{pgfplots}
\DeclareUnicodeCharacter{2212}{−}
\usepgfplotslibrary{groupplots,dateplot}
\usetikzlibrary{patterns,shapes.arrows}
\pgfplotsset{compat=newest}
\usetikzlibrary{arrows}
\usepackage{xcolor}

\newcommand{\obs}{\mathbf{o}}
\newcommand{\act}{\mathbf{a}}
\newcommand{\traj}{\mathbf{\tau}}

\icmltitlerunning{Towards Fusing Point Cloud and Visual Representations for Imitation Learning}

\begin{document}

\twocolumn[
\icmltitle{Towards Fusing Point Cloud and Visual Representations for Imitation Learning}



\icmlsetsymbol{equal}{*}

\begin{icmlauthorlist}
\icmlauthor{Atalay Donat}{equal,kit}
\icmlauthor{Xiaogang Jia}{equal,kit}
\icmlauthor{Xi Huang}{kit}
\icmlauthor{Aleksandar Taranovic}{kit}
\icmlauthor{Denis Blessing}{kit}
\icmlauthor{Ge Li}{kit}
\icmlauthor{Hongyi Zhou}{kit}
\icmlauthor{Hanyi Zhang}{uk}
\icmlauthor{Rudolf Lioutikov}{kit}
\icmlauthor{Gerhard Neumann}{kit}
\end{icmlauthorlist}

\icmlaffiliation{kit}{Karlsruhe Institute of Technology, Germany}
\icmlaffiliation{uk}{University of Liverpool}

\icmlcorrespondingauthor{Xiaogang Jia}{jia266163@gmail.com}
\icmlkeywords{Imitation Learning, Robot Learning}

\vskip 0.3in
]



\printAffiliationsAndNotice{\icmlEqualContribution} 

\begin{abstract}
Learning for manipulation requires using policies that have access to rich sensory information such as point clouds or RGB images. 
Point clouds efficiently capture geometric structures, making them essential for manipulation tasks in imitation learning. In contrast, RGB images provide rich texture and semantic information that can be crucial for certain tasks. 
Existing approaches for fusing both modalities assign 2D image features to point clouds. However, such approaches often lose global contextual information from the original images.
In this work, we propose FPV-Net, a novel imitation learning method that effectively combines the strengths of both point cloud and RGB modalities. Our method conditions the point-cloud encoder on global and local image tokens using adaptive layer norm conditioning, leveraging the beneficial properties of both modalities. 
Through extensive experiments on the challenging RoboCasa benchmark, we demonstrate the limitations of relying on either modality alone and show that our method achieves state-of-the-art performance across all tasks.
\end{abstract}
\section{Introduction}
\label{sec:introduction}


Imitation Learning (IL) has become a fundamental approach in robotic learning \cite{brohan2022rt, chi2023diffusion, zhao2023learning, black2024pi_0, kim24openvla}, allowing agents to acquire complex behaviors by mimicking expert demonstrations. IL can additionally benefit from contextual information that provides task description, therefore reducing the need for inferring task goal from the demonstrations \cite{ding2019goal}. 
A crucial aspect of IL is the choice of the used input representation, as it directly impacts the agent's ability to generalize and make informed decisions. 
RGB images are a common input modality because they offer rich texture and semantic information that can be critical for tasks involving object recognition and contextual reasoning \cite{mandlekar2021matters,reuss2024multimodal,liu2024rdt}.  Additionally, they are easy to obtain and relatively cheap, making them a practical choice in many scenarios.
Another input modality is a point cloud \cite{zhu2024point, ze20243d, ke20243d}, which provides us with geometric information. Point cloud representations have proven highly effective for robotic manipulation due to their ability to directly encode 3D spatial structures. A further modality are language instructions.  They contain relevant task context \cite{stepputtis2020language, li2023vision, reuss2024multimodal}, such as human understandable task descriptions. All these input types provide different benefits and limitations in the learning process, and we should fuse them appropriately to extract all the individual benefits, while offsetting the limitations. Therefore, fusing different modalities is a relevant but challenging problem.

In this paper, we focus on the fusion of RGB images and point clouds while also taking language instructions into account. Despite their complementary nature, integrating these RGB images and point clouds remains a significant challenge in IL. Existing approaches \cite{gervet2023act3d, shridhar2023perceiver, ze20243d} primarily attempt to assign 2D visual features to point clouds, thereby incorporating RGB information into 3D representations. However, such strategies often fail to retain the global contextual information from images, leading to suboptimal performance in tasks that require both precise spatial reasoning and high-level semantic understanding. As a result, neither modality alone—nor naïve fusion techniques—achieves universally strong performance across diverse imitation learning benchmarks. Yet, more recent approaches of combining modalities such as adaptive conditioning in Layer-Norm layers \cite{Peebles2022DiT} has not yet been explored in the imitation learning context, even though it allows a more flexible sensor fusion scheme. 

\begin{figure}[t]
    \centering
    \includegraphics[width=\linewidth]{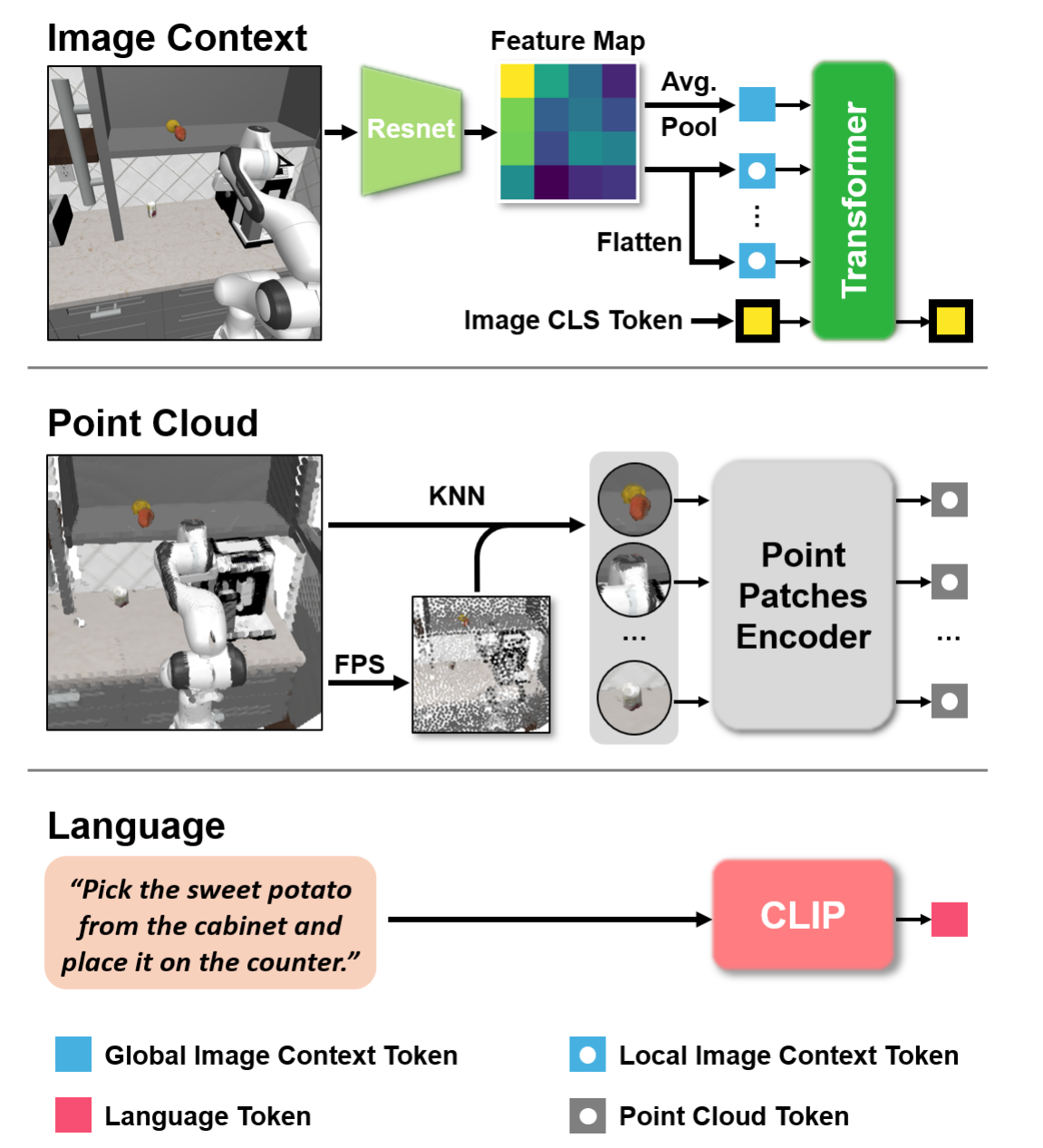}
    \caption{Processing each input modality to generate corresponding embeddings. \textbf{Top}: A FiLM-ResNet  architecture is used to extract a feature map from the context image. The feature map is processed through average pooling and flattening to obtain global and local feature tokens, which are then concatenated and fed into the transformer along with a learnable CLS token, whose output is used as a condition vector for the diffusion policy (Figure \ref{fig:dit_block}). \textbf{Middle}: The point cloud input is processed by applying FPS to sample points, followed by KNN to group point patches using these FPS points as centers. The resulting patches are passed through a point patches encoder, which can be a lightweight MLP or the pretrained SUGAR model. \textbf{Bottom}: The CLIP model is employed to generate the language embedding for the behavior prompt.} 
    \label{fig:input_components}
    \vspace{-0.5cm}
\end{figure}

\begin{figure}[t]
    \centering
    \includegraphics[width=\linewidth]{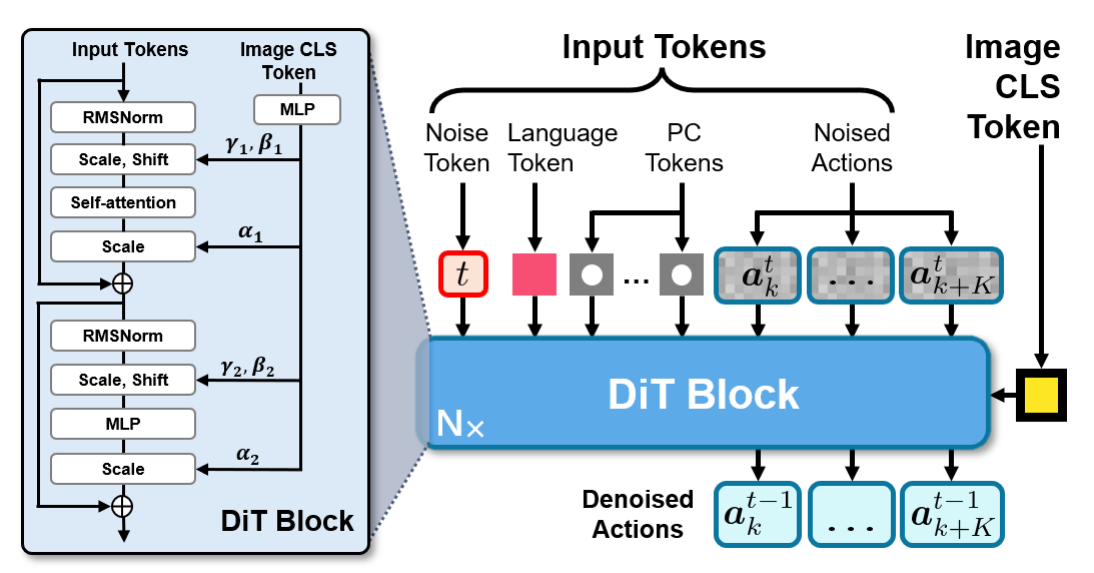}
    \caption{Conditioned on image CLS tokens, the transformer-based diffusion policy (DiT block) denoises action chunk tokens by utilizing 3D point cloud tokens and language tokens as inputs. The conditioning process is detailed within the structure of the DiT block.}
    \label{fig:dit_block}
\end{figure}

To address this limitation, we introduce \textbf{F}usion of \textbf{P}oint Cloud and \textbf{V}isual Representation \textbf{Net}work (FPV-Net), a novel imitation learning method designed to effectively align and balance the strengths of both point cloud and RGB images. Our approach leverages novel conditioning methods for sensor fusion \cite{Peebles2022DiT} and ensures that the geometric precision of point clouds is preserved while leveraging the global semantic richness of RGB inputs, enabling a more robust and generalizable policy learning process.
For the extraction of representations from RGB images, we use a neural network based on the FiLM-ResNet architecture \cite{perez2018film}. This extraction process is conditioned on the language instruction, thus effectively incorporating this modality into our method. Moreover, we make use of both local features and global features, which we show to be critical for the manipulation tasks. 
To extract data from point clouds, we apply Furthest Point Sampling \cite{eldar1994farthest} and k-Nearest Neighbors, that are then encoded into learned embeddings.  For fusing the modalities, we explore 3 different approaches, and show that fusing Point Cloud and Language as main modalities while using RGB images as the conditional modality using AdaLN conditioning \cite{Peebles2022DiT} performs best. Figure \ref{fig:input_components} illustrates how FPV-Net extracts features from different modalities.

We evaluate FPV-Net on RoboCasa \citep{robocasa2024}, a challenging benchmark for robotic manipulation. We conduct extensive experiments to analyze the impact of different input modalities. Our results indicate that neither point clouds nor RGB images alone provide optimal performance across all tasks, whereas naïve fusion methods often degrade performance due to poor alignment between modalities.
FPV-Net consistently outperforms state-of-the-art approaches \cite{ke20243d,ze20243d} across all tasks, establishing a new benchmark in multimodal imitation learning.


    


To summarize, our main contributions are threefold. First, we conduct systematic experiments on RoboCasa, showing that neither RGB images nor point clouds alone are sufficient for strong performance, as each modality excels in some tasks but performs poorly in others. Second, we introduce FPV-Net, a diffusion-based multi-modal imitation learning method that leverages point cloud inputs as the main modality and visual inputs as a conditional modality, integrated via AdaLN conditioning \cite{Peebles2022DiT}, while also incorporating language instructions for contextual guidance. FPV-Net achieves state-of-the-art performance across most tasks, and, to our knowledge, using AdaLN to fuse point cloud and RGB modalities is a novel insight. Third, we demonstrate the critical role of local RGB features in fine-grained robotic manipulation tasks, showing that integrating both global and local features significantly enhances model performance.



\section{Related Works}
\label{sec:related_works}

\textbf{Visual Imitation Learning.}
Recent state-of-the-art imitation learning methods \cite{chi2023diffusion,reuss2024efficient, kim24openvla,liu2024rdt,li2025gr} often use 2D images as state representation due to their rich global information and ease of acquisition from raw sensory inputs. However, 2D images lack explicit 3D information such as precise 3D coordinates and object geometry \cite{zhu2024point}, which are crucial for many robotic manipulation tasks. While using multiple camera views can partially mitigate this drawback, it requires significantly more training data to infer the 3D spatial information effectively \cite{ze20243d}. Moreover, image-based policies struggle with occlusions and viewpoint variations \cite{peri2024point}, making generalization across diverse environments challenging.

\textbf{Imitation Learning with 3D Scene Representation.}
An alternative approach is to leverage 3D scene representations, such as point cloud \cite{zhu2024point, ze20243d, ke20243d}, which provide explicit spatial structure and thus enable better spatial reasoning. However, using point clouds usually requires down-sampling \cite{eldar1994farthest}, leading to loss of fine-grained information from the raw sensory data. 
Recently, several studies \cite{shridhar2023perceiver, gervet2023act3d, ke20243d} have investigated how to effectively incorporate both 2D and 3D representations into imitation learning. For instance, Act3D \cite{gervet2023act3d} generates feature clouds using multi-view RGB images and depth information. 3D Diffuser Actor \cite{ke20243d} lifts ResNet features to 3D using the depth map. Unlike these approaches, FPV-Net introduces a novel 2D-3D fusion strategy by conditioning Transformer policy with 2D images from multiple views while processing tokenized 3D representations, enabling better generalization and spatial reasoning.

\textbf{Multi-modal Sensory Fusion in Imitation Learning.}
Most existing research on multi-modal sensory fusion in imitation learning focuses on combining image observations with language goal conditioning. A common strategy is to treat image and language inputs as separate tokens within a Transformer and train the policy from scratch \cite{reuss2024multimodal, bharadhwaj2024roboagent}. Another line of research leverages large pre-trained Vision-Language Models (VLMs) and fine-tunes them with demonstrations to create Vision-Language-Action (VLA) models \cite{cheang2024gr, kim24openvla, black2024pi_0}. However, these methods predominantly rely on 2D image features, which limits their effectiveness when working with small datasets or tasks requiring detailed spatial reasoning. In the contrary, FPV-Net fuses 2D and 3D observations, enabling more efficient multi-modal learning.

\section{Preliminaries}
\label{sec:preliminaries}

\subsection{Problem Formulation}

Imitation learning (IL) aims to train an agent to perform tasks by learning from expert demonstrations. Given a dataset of expert trajectories \(\mathcal{D} = \{(\traj_i)\}_{i=1}^{N}\), where each trajectory \(\traj_i\) consists of a sequence of observations and corresponding expert actions

\begin{equation}
    \traj_i = (\obs_1, \act_1, \obs_2, \act_2, \dots, \obs_K, \act_K),
\end{equation}

the goal is to learn a policy \(\pi(\act|\obs): \mathcal{O} \to \mathcal{A}\) that maps observations to actions in a manner that mimics expert behavior.





\subsection{Multi-Modal Imitation Learning}
In a multi-modal imitation learning framework, the agent receives a multi-modal observation at each time step \(k\) consisting of:

\textbf{Language instruction} (\(\mathbf{x}_k^L\)): Provides high-level task semantics and contextual guidance, enabling the agent to generalize across diverse instructions.

\textbf{RGB image} (\(\mathbf{x}_k^I\)): Captures visual scene information, including object appearances, spatial arrangements, and environmental semantics.

\textbf{Point cloud} (\(\mathbf{x}_k^P\)): Offers a structured 3D representation of the environment, encoding geometric and spatial relationships that are crucial for manipulation.

Thus, an observation in the framework is defined as 
\begin{equation}
    \obs = (\mathbf{x}_k^L, \mathbf{x}_k^I, \mathbf{x}_k^P) \in \mathcal{O},
\end{equation}
where $\mathcal{O}$ denotes the observation space.
Building on the success of Action Chunking \cite{zhao2023learning} in Imitation Learning, we formulate the objective as predicting a sequence of future actions
\begin{equation}
    \act = (\act_k, \act_{k+1}, \dots, \act_{k+H}) \in \mathcal{A}^H,
\end{equation}
where \(H\) is the prediction horizon, and $\mathcal{A}$ denotes the action space.


\subsection{Score-Based Diffusion Policies}

FPV-Net adopts the continuous-time denoising diffusion model from EDM \cite{karras2022elucidating} to represent the policy. Denoising diffusion models aim to time-reverse a stochastic noising process that transforms the data distribution into Gaussian noise \cite{song2020denoising}, allowing for generating new samples that are distributed according to the data. In FPV-Net, a score-based diffusion model is used for the policy \(\pi(\act|\obs)\). The denoising process is governed by a stochastic differential equation (SDE) given by
\
\begin{equation}
\small
\label{eq: conditional Karras Song SDE}
\begin{split}
\mathrm d \act =  \big( \beta_t \sigma_t - \dot{\sigma}_t  \big) \sigma_t \nabla_\act \log p_t(\act | \obs)  \mathrm dt + \sqrt{2 \beta_t} \sigma_t \mathrm d B_t,
 \end{split}
\end{equation}

where \(\beta_t\) determines how much noise is injected, $B_t$ denotes a standard Wiener process, and \(p_t(\act | \obs)\) is the score function of the diffusion process which moves samples towards regions of high data density. To generate new samples from noise, one trains a neural network to approximate \(\nabla_\act \log p_t(\act | \obs)\) using Score Matching (SM) \cite{6795935}. The SM objective is
\begin{equation}
\label{eq: denoising score matching loss}
\begin{split}
     \mathcal{L}_{D_{\theta}} = 
     \mathbb{E}_{\mathbf{\sigma_t}, \act, \boldsymbol{\epsilon}} \left[  \alpha (\sigma_t) \| D_{\theta}(\act + \boldsymbol{\epsilon}, \obs, \sigma_t)  - \act  \|_2^2 \right],
     \end{split}
\end{equation}
where $D_{\theta}(\act + \boldsymbol{\epsilon}, \obs, \sigma_t)$ is the trainable network. During training, noise is sampled from a predefined distribution and added to an action sequence. The network then predicts the denoised actions and computes the SM loss.
Once training is complete, new action sequences can be generated by starting from random noise and approximating the reverse SDE in discrete steps using a numerical ODE solver. Specifically, one samples an initial action $\act_t \sim \mathcal{N}(0, \sigma_t^2 I)$ from the prior and progressively denoises it. In FPV-Net, this is accomplished via the DDIM-solver \cite{song2020denoising}, which is an ODE solver tailored for diffusion models that can denoise actions in just a few steps. In all experiments, FPV-Net uses 4 denoising steps.

\section{Method}
\label{sec:model}

Fusion of Point Cloud and Visual representation Network (FPV-Net) is a multi-modal transformer-based diffusion policy which leverages point cloud, image and language inputs. In this section, we introduce how we process these different modalities and propose three different fusion methods to combine point cloud features and image features. An overview of our model is shown in Figures \ref{fig:input_components} and \ref{fig:dit_block}.







\subsection{Image Processing}
\label{sec:rgb_processing}
To extract meaningful representations from RGB inputs, we utilize a FiLM-ResNet architecture \cite{perez2018film}, which is conditioned on the language instructions. This approach allows the model to modulate feature extraction based on linguistic context, improving the alignment between vision and language modalities.
Most prior works \cite{chi2023diffusion,zhao2023learning,reuss2024multimodal} in imitation learning extract only a global token from ResNet \cite{he2016deep} feature maps, discarding fine-grained local spatial information. However, we argue that both global and local features are critical for capturing fine-grained visual details necessary for action prediction.
To address this, we extract features as follows:

\textbf{Global Token}: We apply global average pooling over the ResNet feature map to obtain a single global representation.

\textbf{Local Tokens}: Instead of discarding spatial features, we flatten the feature map into a sequence of local tokens, preserving important spatial details.

Finally, we concatenate the global token with the local feature tokens, forming a comprehensive visual representation
\begin{equation}
\small
    \mathbf{z}^I_t = \text{Concat}(\text{AvgPool}(F_{\text{ResNet}}(\mathbf{I})), \text{Flatten}(F_{\text{ResNet}}(\mathbf{I}))),
\end{equation}

where $F_{\text{ResNet}}(\mathbf{I})$ denotes the extracted feature map from FiLM-ResNet.
This enriched representation provides the policy with a multi-scale visual understanding, ensuring that both high-level semantics and fine-grained local details contribute to decision-making. The illustration of the image processing can be found in Figure \ref{fig:input_components}.


\subsection{Point Cloud Processing}
\label{sec:pc_processing}

Prior approaches in 3D imitation learning, such as 3D Diffusion Policy (DP3) \cite{ze20243d} and 3D Diffuser Actor (3DA) \cite{ke20243d}, suffer from key limitations. DP3’s max pooling discards local geometric features, while 3DA’s 2D feature lifting loses global contextual information from original images. Moreover, 3DA generates an excessive number of point tokens, leading to higher computational costs.
To effectively process a point cloud \( \mathbf{x}_t^P \in \mathbb{R}^{N \times 3} \) consisting of \( N \) points in 3D space, we construct a structured representation as follows:

\textbf{Furthest Point Sampling (FPS)} \cite{eldar1994farthest, qi2017pointnet}: We sample \( M = 256 \) center points, ensuring a coverage of the global geometric structure.

\textbf{k-Nearest Neighbors (KNN)} \cite{qi2017pointnet++}: For each center point, we retrieve its \( K = 32 \) nearest neighbors, forming local point groups that capture fine-grained spatial structures.


Each local point group is encoded into a latent representation using a point cloud encoder \( \psi_\theta \). The final point cloud embedding is represented as
\begin{equation}
    \mathbf{z}_t^P = \{ \psi_\theta(\textbf{G}_m) \}_{m=1}^{M}, \quad \mathbf{z}_t^P \in \mathbb{R}^{M \times d},
\end{equation}

where $\textbf{G}_m \in \mathbb{R}^{K \times 3}$ represents the $K$-neighbor subset for the $m$-th sampled center, $\psi_\theta(\cdot)$ is the point cloud encoder that extracts a per-group embedding, and $\mathbf{z}_t^P$ consists of $M = 256$ tokens, each of dimension $d$.
By structuring the point cloud representation into a tokenized format, our approach preserves both local fine-grained features and global contextual information, ensuring a more expressive representation for 3D imitation learning.
We explore two different point cloud encoding strategies:

\textbf{Lightweight MLP Encoder}: Inspired by 3D Diffusion Policy \cite{ze20243d}, we use a multi-layer perceptron (MLP) followed by a max pooling layer to process each point group independently. This method is computationally efficient and preserves local structures.

\textbf{Pretrained SUGAR Model}: We leverage a pretrained point cloud encoder, SUGAR \cite{Chen_2024_SUGAR}, to extract richer and more informative features, benefiting from knowledge gained in large-scale 3D datasets.







\begin{figure*}[h!]
    \centering
    \includegraphics[width=\textwidth]{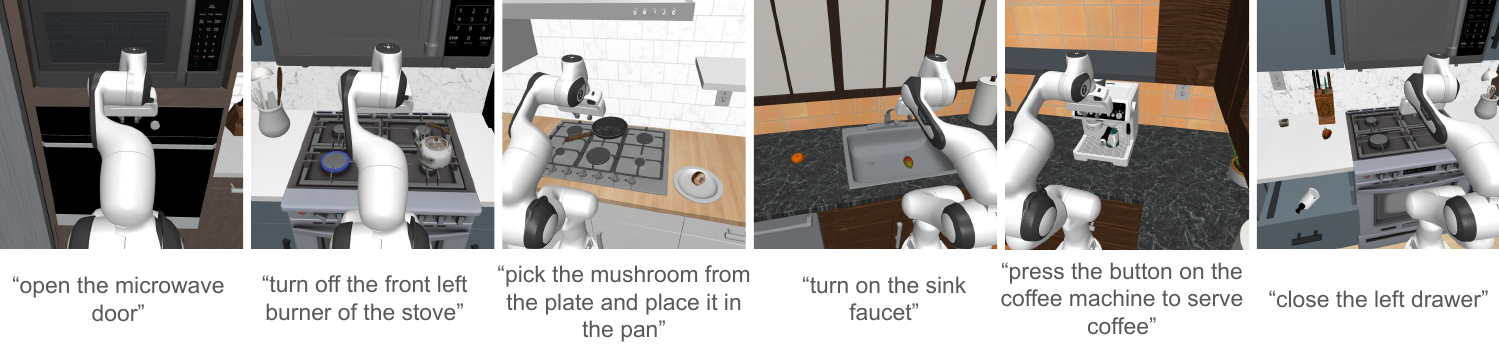}
    \caption{Example scenarios from the RoboCasa benchmark \cite{robocasa2024} used in our experiments.}
    \label{fig:robocasa}
\end{figure*}

\subsection{Fusing Multi-Modal Embeddings}
\label{sec:modality_fusion}

To effectively integrate multi-modal observations, including RGB images, point clouds, and language embeddings, we explore three different fusion strategies for combining image and point cloud features. In the following, other than the image embedding $ \mathbf{z}_t^I$ and the point cloud embedding $\mathbf{z}_t^P$, we use $\mathbf{z}_t^L \in \mathbb{R}^{d_L}$ to denote language embeddings, which are obtained via the frozen CLIP model \cite{radford2021learningtransferablevisualmodels}.

\subsubsection{Concatenation-Based Fusion}
A straightforward approach is to directly concatenate the embeddings of these three modalities and use it as input for the transformer policy. This fused representation $\mathbf{z}_t^{\text{fusion}}$ can be written as

\begin{equation}
    \mathbf{z}_t^{\text{fusion}} = \text{Concat}(\mathbf{z}_t^I, \mathbf{z}_t^P, \mathbf{z}_t^L).
\end{equation}
Although this fusion retains all feature information, it lacks a structured interaction between modalities.

\subsubsection{Adaptive LayerNorm Conditioning}
\label{sec:adaln}
Inspired by the use of Adaptive LayerNorm (AdaLN) layers to condition on classes in DiT models \cite{Peebles2022DiT}, we explore using AdaLN conditioning layers not on language, but on the point cloud or the image modality. In this way, AdaLN conditioning treats one modality as conditioning input and the other modalities as main feature inputs.
The conditioning inputs scale or shift main feature within the attention mechanism

\[
\text{AdaLN}(\mathbf{z} \mid \mathbf{\mathbf{c}}) = \gamma(\mathbf{c}) \odot \frac{\mathbf{z} - \mu(\mathbf{z})}{\sigma(\mathbf{z})} + \beta(\mathbf{c}),
\]

where $\mathbf{z}$ is the main feature, $\mathbf{c}$ is the conditioning input, $\mu(\mathbf{z})$ and $\sigma^2(\mathbf{z})$ are the mean and variance of the main input $\mathbf{z}$, and $\gamma(\mathbf{c})$ and $\beta(\mathbf{c})$ are learnable functions that map the conditioning input to a pair of scale and shift parameters. More details about AdaLN conditioning can be found in Appendix \ref{sec:adaln}.

\paragraph{Image and Language as Main Modality}
In this setup, we select the image embeddings $\mathbf{z}_t^I$ and language embeddings $\mathbf{z}_t^L$ as the primary modality. The AdaLN layers take the point cloud embeddings $\mathbf{z}_t^P$ as conditions to modulate the activation of the primary modality. The fusion is formulated as

\begin{equation}
    \mathbf{z}_t^{\text{fusion}} =  \text{AdaLN}(\mathbf{z}_t^I, \mathbf{z}_t^L | \mathbf{z}_t^P).
\end{equation}

\paragraph{Point Cloud and Language as Main Modality}
Alternatively, we consider using point cloud embedding and language embedding as primary modality and image embedding as conditions

\begin{equation}
    \mathbf{z}_t^{\text{fusion}} =  \text{AdaLN}(\mathbf{z}_t^P, \mathbf{z}_t^L | \mathbf{z}_t^I).
\end{equation}

The observation embedding \(\mathbf{z}_t^{\text{fusion}}\) will then be fed into the transformer-based diffusion policy (Figure \ref{fig:dit_block}).


\section{Experiments}
\label{sec:experiments}

We conduct extensive experiments to answer the following questions:

\textbf{Q1)} Is a single modality enough to perform efficiently on challenging environments?

\textbf{Q2)} How does our method compare with state-of-the-art imitation learning policies?

\textbf{Q3)} What kinds of fusion types are most powerful?
\begin{table*}[t!]
\begin{center}
\begin{sc}
\resizebox{\textwidth}{!}{%
\renewcommand{\arraystretch}{1.3}
\setlength{\aboverulesep}{0pt}
\setlength{\belowrulesep}{0pt}
\begin{tabular}{ll|ccc|ccc|cc}
\toprule
\textbf{Category} & \textbf{Task} & BC & DP3 & 3DA & PC-only & RGB-only & PC+RGB & FPV-MLP & FPV-SUGAR \\
\midrule
\multirow{8}{*}{Pick and Place}
& PnPCabToCounter
& $0.02$
& $0.04$
& $0.00$
& $0.02$
& $0.00$
& $0.04$
& \textbf{\upshape 0.16}
& $0.10$
\\
& PnPCounterToCab
& $0.06$
& $0.02$
& $0.00$
& $0.00$
& $0.00$
& $0.08$
& $0.08$
& \textbf{\upshape 0.14}
\\
& PnPCounterToMicrowave
& $0.02$
& $0.06$
& $0.00$
& $0.00$
& $0.02$
& $0.10$
& \textbf{\upshape 0.26}
& $0.10$
\\
& PnPCounterToSink
& $0.02$
& $0.00$
& $0.00$
& $0.00$
& $0.02$
& $0.04$
& $0.06$
& \textbf{\upshape 0.08}
\\
& PnPCounterToStove
& $0.02$
& $0.00$
& $0.00$
& $0.00$
& $0.00$
& $0.02$
& \textbf{\upshape 0.06}
& $0.04$
\\
& PnPMicrowaveToCounter
& $0.02$
& $0.00$
& $0.00$
& $0.02$
& $0.00$
& $0.04$
& $0.08$
& \textbf{\upshape 0.12}
\\
& PnPSinkToCounter
& $0.08$
& $0.00$
& $0.00$
& $0.00$
& $0.00$
& $0.18$
& $0.22$
& \textbf{\upshape 0.30}
\\
& PnPStoveToCounter
& $0.06$
& $0.00$
& $0.00$
& $0.02$
& $0.02$
& $0.06$
& $0.20$
& \textbf{\upshape 0.26}
\\
\midrule
\multirow{4}{*}{Open/Close Doors}
& OpenSingleDoor
& $0.46$
& $0.24$
& $0.00$
& $0.44$
& $0.38$
& $0.72$
& $0.68$
& \textbf{\upshape 0.74}
\\
& OpenDoubleDoor
& $0.28$
& $0.20$
& $0.00$
& $0.38$
& $0.50$
& $0.86$
& \textbf{\upshape 0.94}
& $0.92$
\\
& CloseDoubleDoor
& $0.28$
& $0.56$
& $0.00$
& $0.50$
& $0.50$
& $0.76$
& \textbf{\upshape 0.82}
& $0.78$
\\
& CloseSingleDoor
& $0.56$
& $0.62$
& $0.14$
& $0.76$
& $0.82$
& $0.80$
& \textbf{\upshape 0.86}
& $0.84$
\\
\midrule
\multirow{2}{*}{Open/Close Drawers}
& OpenDrawer
& $0.42$
& $0.36$
& $0.00$
& $0.36$
& $0.34$
& $0.56$
& $0.62$
& \textbf{\upshape 0.72}
\\
& CloseDrawer
& $0.80$
& $0.48$
& $0.00$
& $0.90$
& $0.94$
& \textbf{\upshape 0.96}
& $0.90$
& $0.94$
\\
\midrule
\multirow{2}{*}{Twisting Knobs}
& TurnOnStove
& $0.32$
& $0.24$
& $0.10$
& $0.48$
& $0.30$
& $0.50$
& $0.46$
& \textbf{\upshape 0.66}
\\
& TurnOffStove
& $0.04$
& $0.06$
& $0.02$
& $0.12$
& $0.10$
& $0.16$
& $0.12$
& \textbf{\upshape 0.20}
\\
\midrule
\multirow{3}{*}{Turning Levers}
& TurnOnSinkFaucet
& $0.38$
& $0.32$
& $0.06$
& $0.40$
& $0.38$
& $0.24$
& $0.68$
& \textbf{\upshape 0.70}
\\
& TurnOffSinkFaucet
& $0.50$
& $0.42$
& $0.28$
& $0.58$
& $0.42$
& $0.34$
& \textbf{\upshape 0.82}
& $0.78$
\\
& TurnSinkSpout
& $0.54$
& $0.54$
& $0.26$
& \textbf{\upshape 0.70}
& $0.48$
& $0.40$
& $0.54$
& $0.52$
\\
\midrule
\multirow{3}{*}{Pressing Buttons}
& CoffeePressButton
& $0.48$
& $0.16$
& $0.08$
& $0.08$
& $0.76$
& $0.86$
& $0.86$
& \textbf{\upshape 0.90}
\\
& TurnOnMicrowave
& $0.62$
& $0.38$
& $0.06$
& $0.24$
& $0.32$
& $0.64$
& \textbf{\upshape 0.74}
& $0.68$
\\
& TurnOffMicrowave
& $0.70$
& $0.54$
& $0.32$
& $0.56$
& $0.66$
& $0.82$
& $0.86$
& \textbf{\upshape 0.96}
\\
\midrule
\multirow{2}{*}{Insertion}
& CoffeeServeMug
& $0.22$
& $0.18$
& $0.00$
& $0.16$
& $0.22$
& $0.42$
& \textbf{\upshape 0.62}
& $0.48$
\\
& CoffeeSetupMug
& $0.00$
& $0.04$
& $0.00$
& $0.00$
& $0.02$
& $0.10$
& \textbf{\upshape 0.22}
& $0.16$
\\
\midrule
\multicolumn{2}{c|}{\textbf{Average Success Rate}}
& $0.2880$
& $0.2275$
& $0.0550$
& $0.2800$
& $0.3000$
& $0.4042$
& $0.4942$
& \textbf{\upshape 0.5050}
\\
\bottomrule
\end{tabular}
}
\end{sc}
\end{center}
\caption{Results for each task in RoboCasa. The models were trained for 100 epochs with 50 human demonstrations per task and evaluated with 50 episodes for each task. The bold numbers highlight the best achieved success rate for that task among all the models. }
\label{table:main_results_2d}
\end{table*}


\subsection{Simulations}

\textbf{RoboCasa} \cite{robocasa2024}:
RoboCasa is a large-scale simulation framework designed to train generalist robots in diverse and realistic household environments, with a particular emphasis on complex kitchen tasks. It features 120 meticulously crafted kitchen scenes, over 2,500 high-quality 3D objects across 150 categories, and 100 tasks divided into foundational atomic tasks and intricate composite tasks. Leveraging generative AI tools, RoboCasa achieves unparalleled diversity, realism, and scalability in robotic learning.
This benchmark is characterized by its exceptional difficulty, stemming from the highly diverse scenarios it presents. Each scenario is accompanied by only one demonstration, significantly increasing the challenge for learning algorithms. For instance, in pick-and-place tasks, the object to be manipulated varies across scenarios, with just one demonstration per case. Furthermore, the training and evaluation datasets feature completely distinct scenes, further testing a model’s ability to generalize and adapt robot behaviors to novel scenarios.
With its extensive task set, environmental variability, and high-fidelity simulations, RoboCasa establishes itself as a new standard for evaluating robotic learning methodologies, pushing the boundaries of generalization and adaptability in robot learning.

\textbf{Training and Evaluation}: We train each method for 100 epochs and rollout the models for 50 times for all tasks in RoboCasa. We group similar tasks together as shown in Table \ref{tab:task_groups} and train the models for each of the groups. 

\subsection{Baselines}

\textbf{BC} \cite{robocasa2024}: We inherit the result reported in RoboCasa. RoboCasa uses the BC-Transformer implemented by RoboMimic. The BC policy uses a CLIP model to encode the goal instruction and a ResNet-18 with FilM layers to encode the image-based observations.

\textbf{3D Diffusion Policy (DP3)} \cite{ze20243d}: DP3
extracts point-wise features from single-view points clouds with a MLP-based encoder and forms a compact 3D visual representation. Robot actions are then generated by a convolutional network-based architecture, conditioned on this representation and the current robot states.

\textbf{3D Diffuser Actor (3DA)} \cite{ke20243d}: 3DA is a diffusion-based policy conditioned on 3D scene features and language instructions.
The 3D scene features are extracted and aggregated from single or multi-view images and depth maps. The policy denoises rotation and translation of the robot's end-effector as action.

\subsection{FPV-Net}
\label{sec:proposed_models}

We systematically evaluate how the FPV-Net deals with different modalities while maintaining a consistent architecture and diffusion policy configuration across all experiments. This setup allows us to directly compare the effectiveness of different representations.

\textbf{PC-only}:
We first group the point cloud by selecting 256 centers via Furthest Point Sampling (FPS), then retrieve 32 nearest neighbors using K-Nearest Neighbors (KNN) to form 256 point groups. Each group is passed through a lightweight MLP encoder, obtaining an embedding per group. These embeddings, along with a language embedding from CLIP, a timestep embedding, and the noisy action, are provided to a transformer-based diffusion policy.

\textbf{RGB-only}:
In this model, each camera view is processed by a ResNet-18 model, which is pretrained and then finetuned separately for each view. FiLM layers condition the network on the CLIP-encoded language instruction. The resulting embeddings from all camera views are subsequently given to the same transformer-based diffusion policy employed in the PC-only model.
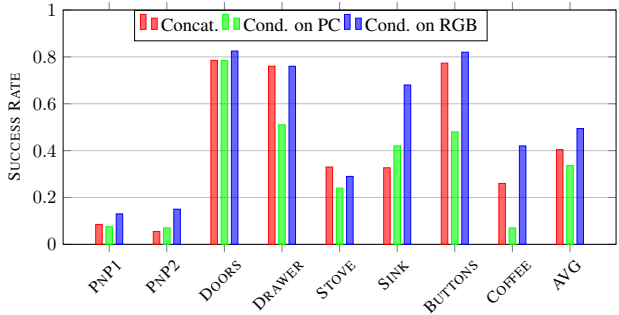
\begin{figure}
    \centering
    \resizebox{0.48\textwidth}{!}{%
        \begin{tikzpicture}[baseline]
        \definecolor{lightgray204}{RGB}{204,204,204}
        \definecolor{crimson}{RGB}{214,39,40}
        \definecolor{darkgray}{RGB}{176,176,176}
        \definecolor{darkorange}{RGB}{255,127,14}
        \definecolor{darkturquoise}{RGB}{23,190,207}
        \definecolor{forestgreen}{RGB}{44,160,44}
        \definecolor{goldenrod}{RGB}{188,189,34}
        \definecolor{gray}{RGB}{127,127,127}
        \definecolor{mediumpurple}{RGB}{148,103,189}
        \definecolor{orchid}{RGB}{227,119,194}
        \definecolor{sienna}{RGB}{140,86,75}
        \definecolor{steelblue}{RGB}{31,119,180}

        \begin{axis}[
            ybar,
            height=6cm, width=12cm,
            bar width=3.5pt,
            ylabel=Success Rate,
            ylabel style={
                font=\footnotesize\scshape
            },
            ymin=0, ymax=1,
            legend style={at={(0.448, 1)}, anchor=north, legend columns=-1},
            symbolic x coords={PnP1, PnP2, Doors, Drawer, Stove, Sink, Buttons, Coffee, AVG},
            xtick=data,
            xticklabel style={
                rotate=45,
                font=\footnotesize\scshape,
                yshift=3pt,
            },
            ymajorgrids
        ]
        
        \addplot [
            draw=red,
            fill=red,
            line width=.1mm,
            fill opacity=0.6
        ] coordinates {
            (PnP1, 0.0850)
            (PnP2, 0.0550)
            (Doors, 0.7850)
            (Drawer, 0.7600)
            (Stove, 0.3300)
            (Sink, 0.3267)
            (Buttons, 0.7733)
            (Coffee, 0.2600)
            (AVG, 0.4042)
        };

        \addplot [
            draw=green,
            fill=green,
            line width=.1mm,
            fill opacity=0.6
        ] coordinates {
            (PnP1, 0.0750)
            (PnP2, 0.0700)
            (Doors, 0.7850)
            (Drawer, 0.5100)
            (Stove, 0.2400)
            (Sink, 0.4200)
            (Buttons, 0.4800)
            (Coffee, 0.0700)
            (AVG, 0.3358)
        };

        \addplot [
            draw=blue,
            fill=blue,
            line width=.1mm,
            fill opacity=0.6
        ] coordinates {
            (PnP1, 0.1300)
            (PnP2, 0.1500)
            (Doors, 0.8250)
            (Drawer, 0.7600)
            (Stove, 0.2900)
            (Sink, 0.6800)
            (Buttons, 0.8200)
            (Coffee, 0.4200)
            (AVG, 0.4942)
        };

        \legend{Concat., Cond. on PC, Cond. on RGB}
        \end{axis}
        \end{tikzpicture}

    }
    \caption{Success rates using different fusion types for point cloud and RGB images.}
    \label{fig:ablation_fusion}
\end{figure}

\textbf{PC+RGB}:
This variant simply concatenates the point group embeddings from PC-only with the RGB embeddings from RGB-only, and feeds the combined representation into the transformer-based diffusion policy.

\textbf{FPV-MLP}:
Here, the point cloud is processed as before, but we additionally exploit local RGB features. Specifically, we use the 8x8 feature map produced by the third ResNet layer for each image. This feature map is flattened and concatenated with the global ResNet embedding, producing 65 tokens per view. Tokens from all views, along with a learnable class token, are passed to a transformer. The output of the class token serves as the condition vector for AdaLN, while the point group embeddings enter the diffusion policy in the usual way. 

\textbf{FPV-SUGAR}:
In this model, we use the point cloud encoder of the pretrained 3D visual representation model SUGAR \cite{Chen_2024_SUGAR}, which also partitions points into 256 groups of 32 via FPS and KNN, but subsequently also employs a 12-layer transformer. We use the model pretrained on multi-object scenes using objects from the Objaverse \cite{objaverse} dataset. To reduce computational cost, we freeze the first 10 layers and finetune only the last 2. The RGB images are processed similarly to FPV-MLP, except that we use the 4x4 feature map from the fourth ResNet layer. Finally, the conditioned transformer-based diffusion policy is applied as before.

\subsection{Main Results}

Table \ref{table:main_results_2d} shows that models utilizing both modalities outperform those using a single modality, which addresses Q1. Simply concatenating point cloud and RGB features leads to a 10\% improvement, illustrating the complementary nature of spatial and semantic information: each modality contributes unique advantages that are not fully captured by the other. Notably, pick-and-place and insertion tasks benefit most from having both modalities, suggesting that both spatial and semantic cues are crucial for manipulating objects unseen during training.
In one particular task the PC-only method performs noticeably better than the other models, namely the \textsc{TurnSinkSpout} task, which requires further investigation.
\begin{figure}
    \centering
    \resizebox{0.48\textwidth}{!}{
    \begin{tikzpicture}[baseline]
    \definecolor{lightgray204}{RGB}{204,204,204}
    \definecolor{crimson}{RGB}{214,39,40}
    \definecolor{darkgray}{RGB}{176,176,176}
    \definecolor{darkorange}{RGB}{255,127,14}
    \definecolor{darkturquoise}{RGB}{23,190,207}
    \definecolor{forestgreen}{RGB}{44,160,44}
    \definecolor{goldenrod}{RGB}{188,189,34}
    \definecolor{gray}{RGB}{127,127,127}
    \definecolor{mediumpurple}{RGB}{148,103,189}
    \definecolor{orchid}{RGB}{227,119,194}
    \definecolor{sienna}{RGB}{140,86,75}
    \definecolor{steelblue}{RGB}{31,119,180}
    
    \begin{axis}[
        height=6cm,
        width=12cm,
        bar width=3.5pt,
        ylabel=Success Rate,
        ylabel style={
            font=\footnotesize\scshape
        },
        ymin=0, ymax=1,
        legend style={at={(0.348, 1)}, anchor=north, legend columns=-1},
        symbolic x coords={drawer, stove, sink, buttons, coffee},
        xtick=data,
        xticklabels={Drawer,Stove,Sink,Buttons,Coffee},
        xticklabel style={font=\scshape},
        ymajorgrids,
        grid=both,
    ]
    
    \addplot [
        draw=red,
        line width=1pt,
        mark=x,
        mark options={solid},
    ]
    coordinates {
        (drawer, 0.5500)
        (stove, 0.2000)
        (sink, 0.3800)
        (buttons, 0.5200)
        (coffee, 0.0600)
    };
    
    \addplot [
        draw=blue,
        line width=1pt,
        mark=x,
        mark options={solid},
    ]
    coordinates {
        (drawer, 0.6400)
        (stove, 0.2400)
        (sink, 0.5133)
        (buttons, 0.6533)
        (coffee, 0.2900)
    };
    
    \legend{Max Pooling, Transformer}
    \end{axis}
    
    \end{tikzpicture}
    }
    \caption{Success rates using max pool or transformer to obtain global feature vector of RGB images to use in AdaLN conditioning.}
    \label{fig:ablation_cond_vec}
\end{figure}
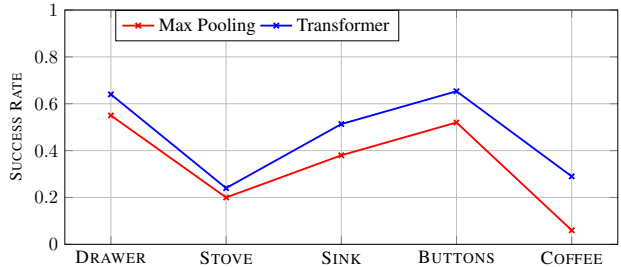

Our PC-only approach outperforms 3D Diffusion Policy by a margin of 5.25\%, answering Q2. A likely explanation is that the max-pooling step discards spatial information critical to the diffusion policy. By contrast, our approach retains more of the point cloud’s geometric structure. Furthermore, grouping points instead of handling each point separately like DP3 allows our PC-only model to better capture local spatial features.

FPV-MLP and FPV-SUGAR, conditioning on RGB features, offer further gains, yielding an average success rate of around 50\%, higher than the simple concatenation of modalities. This suggests the diffusion policy exploits the rich texture and semantic details from RGB data when using AdaLN for conditioning more effectively than taking these features purely as an additional input. Another possible reason is that the transformer-based diffusion policy can better separate the two modalities, focusing on spatial relations through self-attention over point groups while annotating each group with semantic features via AdaLN conditioning.

3DA exhibits a very low success rate on RoboCasa in our experiments. This may be attributed to our decision to train each model for 100 epochs to ensure a fair comparison. However, as a relatively more complex model, 3DA likely requires a longer training duration to achieve optimal performance.

\subsection{Ablation on different fusion}

We compared the performance of different fusion strategies for integrating point cloud and RGB embeddings within the transformer architecture. Concat. refers to a straightforward concatenation of both embeddings. Cond. on PC denotes using RGB features as the main modality while conditioning on point cloud features through AdaLN conditioning. Conversely, Cond. on RGB treats point cloud features as the primary modality, with RGB features providing the conditioning signal via AdaLN. As shown in Figure \ref{fig:ablation_fusion}, conditioning the RGB-based transformer on point cloud features underperforms compared to simple concatenation. This could be due to compressing the entire point cloud into a single vector, which may discard crucial spatial details, particularly for tasks like \textsc{Coffee}, where precise grasping of a mug is required. In contrast, conditioning on RGB features yields the best performance across most tasks, effectively addressing Q3.


\newcommand{\D}{8} 
\newcommand{\U}{10} 

\newdimen\R 
\R=3.5cm 
\newdimen\L 
\L=4cm
\newdimen\B
\B=4.2cm
\newdimen\C
\C=3.8cm

\newcommand{\A}{360/\D} 

\begin{figure}
    \centering
    \includegraphics[width=0.4\textwidth]{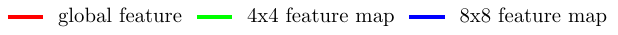}
    \begin{sc}
    \resizebox{0.4\textwidth}{!}{ 
        \begin{tikzpicture}[scale=1]
            \path (0:0cm) coordinate (O); 

            \foreach \X in {1,...,\D}{
                \draw (\X*\A:0) -- (\X*\A:\R);
            }

            \foreach \Y in {0,...,\U}{
                \foreach \X in {1,...,\D}{
                    \path (\X*\A:\Y*\R/\U) coordinate (D\X-\Y);
                }
                \draw [opacity=0.3] (0:\Y*\R/\U) \foreach \X in {1,...,\D}{
                    -- (\X*\A:\Y*\R/\U)
                } -- cycle;
            }

            \path (1*\A:\L) node (L1) {PnP1};
            \path (2*\A:\C) node (L2) {PnP2};
            \path (3*\A:\L) node (L3) {Doors};
            \path (4*\A:\B) node (L4) {Drawer};
            \path (5*\A:\L) node (L5) {Stove};
            \path (6*\A:\C) node (L6) {Sink};
            \path (7*\A:\L) node (L7) {Buttons};
            \path (8*\A:\B) node (L8) {Coffee};

            \draw [color=red,line width=1.5pt]
                ($(D1-1)!0.5!(D1-2)$) --
                ($(D2-1)!0.7!(D2-2)$) --
                ($(D3-8)!0.45!(D3-9)$) --
                ($(D4-6)!0.4!(D4-7)$) --
                ($(D5-2)!0.4!(D5-3)$) --
                ($(D6-5)!0.133!(D6-6)$) --
                ($(D7-6)!0.533!(D7-7)$) -- cycle;

            \draw [color=green,line width=1.5pt]
                ($(D1-1)!0.4!(D1-2)$) --
                ($(D2-1)!0.35!(D2-2)$) --
                ($(D3-7)!0.85!(D3-8)$) --
                ($(D4-7)!0.8!(D4-8)$) --
                ($(D5-3)!0.5!(D5-4)$) --
                ($(D6-4)!0.667!(D6-5)$) --
                ($(D7-8)!0.667!(D7-9)$) --
                ($(D8-3)!0.4!(D8-4)$) -- cycle;

            \draw [color=blue,line width=1.5pt]
                ($(D1-1)!0.3!(D1-2)$) --
                ($(D2-1)!0.5!(D2-2)$) --
                ($(D3-8)!0.25!(D3-9)$) --
                ($(D4-7)!0.6!(D4-8)$) --
                ($(D5-2)!0.9!(D5-3)$) --
                ($(D6-6)!0.8!(D6-7)$) --
                ($(D7-8)!0.2!(D7-9)$) --
                ($(D8-4)!0.2!(D8-5)$) -- cycle;

            \draw[<->, very thick, draw=brown, text=brown]
                ($(D1-7)!0!(D1-8)$)
             -- ($(D1-8)!0!(D1-9)$)
                node[midway, above, yshift=4.5pt, xshift=-1.5pt, inner sep=1pt]
                {\textbf{0.10}};

        \end{tikzpicture}
    }
    \end{sc}
    \caption{Success rates of conditioning on ResNet features with different granularity level. Each level in the chart corresponds to a 10\% difference in success rate.}
    \label{fig:ablation_rgb_features}
    \vspace{-0.3cm}
\end{figure}
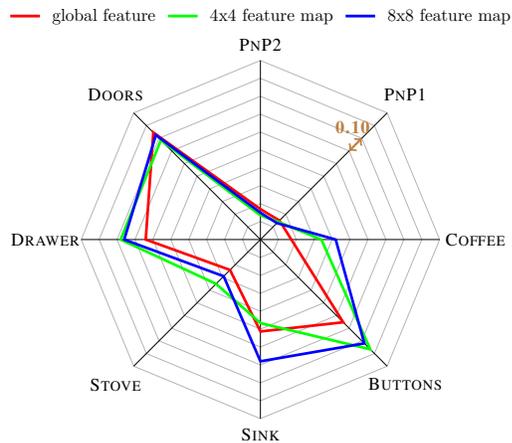

\subsection{Ablation on obtaining condition vector}

AdaLN does not directly support sequences as input, so a single token must be extracted to condition on point clouds or RGB features. We compare two methods: (1) a simple max-pooling layer, and (2) a transformer whose learnable class token serves as the global representation. Figure \ref{fig:ablation_cond_vec} indicates that the transformer-based approach consistently outperforms max pooling in all tested tasks.

\subsection{Ablation on RGB features}
In order to identify the influence of global tokens and local tokens from ResNet feature map, we evaluate FPV-Net with different feature granularity: global features versus 4x4 or 8x8 feature maps. The results are presented in Figure \ref{fig:ablation_rgb_features}, which show that by adding local features from ResNet would gain performance significantly on most tasks such as \textsc{Buttons} and \textsc{Drawers}, whereas the \textsc{Doors} task show less sensitivity. This contrast could be due to the smaller size of buttons and drawer handles, which require finer-grained feature maps for accurate manipulation.


\subsection{Ablation on finetuning SUGAR}

\begin{figure}
    \centering
    \resizebox{0.48\textwidth}{!}{%
        \begin{tikzpicture}[baseline]
        
        \begin{axis}[
            ybar,
            height=6cm, width=12cm,
            bar width=3.5pt,
            ylabel=Success Rate,
            ylabel style={
                font=\footnotesize\scshape
            },
            ymin=0, ymax=1,
            legend style={at={(0.748, 1)}, anchor=north, legend columns=-1},
            symbolic x coords={PnP1, PnP2, Doors, Drawer, Stove, Sink, Buttons, Coffee, AVG},
            xtick=data,
            xticklabel style={
                rotate=45,
                font=\footnotesize\scshape,
                yshift=3pt,
            },
            ymajorgrids
        ]
        
        \addplot [
            draw=red,
            fill=red,
            line width=.2mm,
            fill opacity=0.7
        ] coordinates {
            (PnP1, 0.0400)
            (PnP2, 0.0200)
            (Doors, 0.6250)
            (Drawer, 0.8600)
            (Stove, 0.4200)
            (Sink, 0.6467)
            (Buttons, 0.7867)
            (Coffee, 0.3500)
            (AVG, 0.4292)
        };

        \addplot [
            draw=blue,
            fill=blue,
            line width=.2mm,
            fill opacity=0.7
        ] coordinates {
            (PnP1, 0.0450)
            (PnP2, 0.0500)
            (Doors, 0.6850)
            (Drawer, 0.9000)
            (Stove, 0.3400)
            (Sink, 0.6667)
            (Buttons, 0.8067)
            (Coffee, 0.3700)
            (AVG, 0.4483)
        };

        \legend{No FT, FT 2 layers}
        \end{axis}
        
        \end{tikzpicture}
    }
    \caption{Success rates using different finetuning strategies for the pretrained SUGAR encoder.}
    \vspace{-10pt} 
    \label{fig:ablation_sugar_ft}
\end{figure}
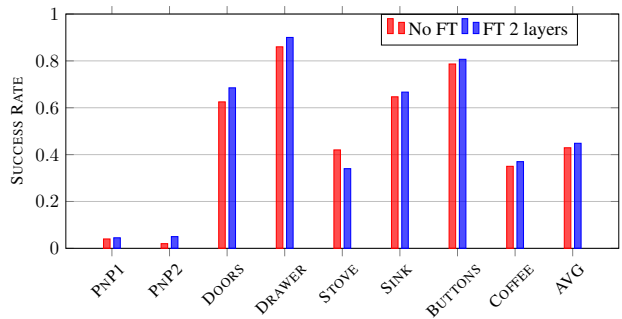

Finally, we examine the effect of different finetuning strategies on FPV-SUGAR. Figure \ref{fig:ablation_sugar_ft} compares a fully frozen SUGAR encoder with an encoder in which only the last two layers are finetuned. With the exception of the stove task, finetuning the last two layers improves performance in nearly every scenario, providing a 2\% boost in average success rate. Finetuning even more layers could potentially further increase the performance of the model.

\section{Conclusion}
\label{sec:conclusion}
In this paper, we introduce the Fusion of Point Cloud and Visual representation Network, a novel approach that integrates RGB and point cloud features using AdaLN conditioning within a transformer. By fusing features at each residual connection, our method effectively captures complementary information from both modalities. Extensive experiments on the RoboCasa benchmark demonstrate significant performance gains over baselines, highlighting the importance of thoughtful cross-modal integration. These results open new avenues for exploring advanced fusion strategies to further enhance robotic perception and understanding of complex environments.



\bibliography{icml2025}
\bibliographystyle{icml2025}

\appendix
\onecolumn

\section{Experiment Settings}
\label{appendix:exp_settings}

RoboCasa is a state-of-the-art simulation framework developed to advance the training of generalist robots in diverse and realistic household settings, particularly in kitchen environments. It comprises 120 meticulously modeled kitchen layouts, over 2,500 high-quality 3D objects spanning 150 categories, and 25 foundational atomic tasks that are the building blocks for robot learning. These atomic tasks encompass essential sensorimotor skills, including pick-and-place, opening and closing doors or drawers, twisting knobs, turning levers, pressing buttons, performing insertions, and navigating kitchen spaces. In our work, we evaluated our model in 24 of these tasks, except for the navigation. A list of these tasks evaluated in our work is given in Table \ref{tab:robo_casa}.

The benchmark is particularly challenging due to its unparalleled diversity and realism. Each scenario includes unique configurations and employs just a single demonstration, significantly raising the bar for generalization. For example, in pick-and-place tasks, the objects vary extensively between scenarios, with no repetitions, forcing models to adapt to new instances without direct prior exposure. Furthermore, the training and evaluation environments are entirely distinct, compelling robotic agents to exhibit robust transfer learning capabilities across unseen kitchens and objects.

These features create a demanding benchmark, testing models on their ability to understand and generalize robotic behavior in highly diverse, real-world-inspired scenarios. RoboCasa's emphasis on realistic physics, photorealistic rendering, and the integration of generative AI tools for diverse asset creation ensures it sets a new standard for evaluating robotic learning methodologies. Its extensive task variability and high fidelity make it one of the most rigorous and comprehensive platforms for advancing generalist robot capabilities in everyday household environments.

\begin{table}[t!]
\renewcommand{\arraystretch}{1.3}
\setlength{\aboverulesep}{0pt}
\setlength{\belowrulesep}{0pt}
\small
    \centering
    \begin{tabularx}{\textwidth}{l l X}
        \toprule
        \textbf{Task}              & \textbf{Skill Family} & \textbf{Description} \\ \hline
        PickPlaceCounterToCabinet   & Pick and place & Pick an object from the counter and place it inside the cabinet. The cabinet is already open. \\ 
        PickPlaceCabinetToCounter   & Pick and place & Pick an object from the cabinet and place it on the counter. The cabinet is already open. \\ 
        PickPlaceCounterToSink      & Pick and place & Pick an object from the counter and place it in the sink. \\\\ 
        PickPlaceSinkToCounter      & Pick and place & Pick an object from the sink and place it on the counter area next to the sink. \\ 
        PickPlaceCounterToMicrowave & Pick and place & Pick an object from the counter and place it inside the microwave. The microwave door is already open. \\ 
        PickPlaceMicrowaveToCounter & Pick and place & Pick an object from inside the microwave and place it on the counter. The microwave door is already open. \\ 
        PickPlaceCounterToStove     & Pick and place & Pick an object from the counter and place it in a pan or pot on the stove. \\ 
        PickPlaceStoveToCounter     & Pick and place & Pick an object from the stove (via a pan or pot) and place it on (the plate on) the counter. \\ \hline
        \\
        OpenSingleDoor              & Opening and closing doors & Open a microwave door or a cabinet with a single door. \\\\
        CloseSingleDoor             & Opening and closing doors & Close a microwave door or a cabinet with a single door. \\\\
        OpenDoubleDoor              & Opening and closing doors & Open a cabinet with two opposite-facing doors. \\ \\
        CloseDoubleDoor             & Opening and closing doors & Close a cabinet with two opposite-facing doors. \\ \\
        OpenDrawer                  & Opening and closing drawers & Open a drawer. \\ \\
        CloseDrawer                 & Opening and closing drawers & Close a drawer. \\\hline
        \\
        TurnOnStove                 & Twisting knobs & Turn on a specified stove burner by twisting the respective stove knob. \\
        TurnOffStove                & Twisting knobs & Turn off a specified stove burner by twisting the respective stove knob. \\ \hline \\
        TurnOnSinkFaucet            & Turning levers & Turn on the sink faucet to begin the flow of water. \\ \\
        TurnOffSinkFaucet           & Turning levers & Turn off the sink faucet to stop the flow of water. \\ \\
        TurnSinkSpout               & Turning levers & Turn the sink spout. \\ \hline \\
        CoffeePressButton           & Pressing buttons & Press the button on the coffee machine to pour coffee into the mug. \\ 
        TurnOnMicrowave             & Pressing buttons & Turn on the microwave by pressing the start button. \\ 
        TurnOffMicrowave            & Pressing buttons & Turn off the microwave by pressing the stop button. \\ \hline \\
        CoffeeSetupMug              & Insertion & Pick the mug from the counter and insert it onto the coffee machine mug holder area. \\
        CoffeeServeMug              & Insertion & Remove the mug from the coffee machine mug holder and place it on the counter. \\
        \bottomrule
    \end{tabularx}
    \caption{The 24 manipulation tasks of RoboCasa used in our work. This list was originally reported as Fig. 11 in \citet{robocasa2024}.}
    \label{tab:robo_casa}
\end{table}

\begin{longtable}{|p{3cm}|p{4cm}|p{8cm}|}
\hline

\end{longtable}
\begin{table}
    \begin{center}    
    \begin{sc}
    \centering
    \setlength{\aboverulesep}{0pt}
    \setlength{\belowrulesep}{0pt}
    \renewcommand{\arraystretch}{1.3}  
    \setlength{\tabcolsep}{6pt}  
    \resizebox{0.7\textwidth}{!}{  
        \begin{tabular}{l l l l}
            \toprule
            \textbf{PnP1} & \textbf{PnP2} & \textbf{Doors} & \textbf{Drawer} \\
            \midrule
            PnPCounterToCab & PnPCounterToMicrowave & OpenSingleDoor & CloseDrawer \\
            PnPCabToCounter & PnPMicrowaveToCounter & CloseSingleDoor & OpenDrawer \\
            PnPCounterToSink & PnPStoveToCounter & OpenDoubleDoor &  \\
            PnPSinkToCounter & PnPCounterToStove & CloseDoubleDoor &  \\
            \midrule
            \textbf{Stove} & \textbf{Sink} & \textbf{Buttons} & \textbf{Coffee} \\
            \midrule
            TurnOnStove & TurnOnSinkFaucet & CoffeePressButton & CoffeeSetupMug \\
            TurnOffStove & TurnOffSinkFaucet & TurnOffMicrowave & CoffeeServeMug \\
             & TurnSinkSpout & TurnOnMicrowave &  \\
            \bottomrule
        \end{tabular}
    }
    \end{sc}
    \end{center}
    \caption{Task groups used for training the models.}
    \label{tab:task_groups}
\end{table}
\section{Hyperparameters}

\begin{table}[h!]
    \centering
    \renewcommand{\arraystretch}{1.4}
    \setlength{\aboverulesep}{0pt}
    \setlength{\belowrulesep}{0pt}
    \begin{adjustbox}{max width=0.8\textwidth}
    \begin{tabular}{l c c c c c c c c c}
        \toprule
        \multirow{2}{*}{\textbf{Hyper-params.}}
        & \multirow{2}{*}{\textbf{PC Only}} 
        & \multirow{2}{*}{\textbf{RGB Only}} 
        & \multirow{2}{*}{\textbf{PC + RGB}}
        & \textbf{PC Cond.}
        & \textbf{RGB Cond.}
        & \textbf{PC Cond. on}\\        
                          &          &          &          &   \textbf{on RGB}    &    \textbf{on PC}    &  \textbf{local RGB feat.}      \\
        \midrule
        Epoch             &     100          &    100      &              100       &          100            &         100             &               100                    \\
        Batch size        &   256             &    256      &              256        &      256                &       256               &       256                            \\
        Optimizer         &    AdamW          &    AdamW      &          AdamW            &           AdamW           &          AdamW            &        AdamW                           \\
        Learning Rate     &         $1\mathrm{e}^{-4}$      &    $1\mathrm{e}^{-4}$      &            $1\mathrm{e}^{-4}$          &           $1\mathrm{e}^{-4}$           &                      $1\mathrm{e}^{-4}$    & $1\mathrm{e}^{-4}$                        \\
        Weight Decay            &    $5\mathrm{e}^{-2}$        &      $5\mathrm{e}^{-2}$      &         $5\mathrm{e}^{-2}$               &        $5\mathrm{e}^{-2}$                &          $5\mathrm{e}^{-2}$              &      $5\mathrm{e}^{-2}$                               \\
        Clip Grad         &          &                   &                      &                      &                      &                                   \\
        Point Sampling    &    FPS      &      -    &    FPS         &          FPS            &          FPS            &          FPS                         \\
        \# Points  &    4096     &     -     &    4096           &       4096               &      4096                &          4096                         \\
        \# Point Groups &    256     &     -     &    256           &       256               &      256                &          256                         \\
        Size of Point Group  &    32     &     -     &    32           &       32               &      32                &          32                         \\
        Latent Dim.      &     512     &                   &                      &                      &                      &                                   \\
        Embedding Dim.      &     128     &    256      &    128      &                                 128           &         256             &      128                             \\
        \bottomrule
    \end{tabular}
    \end{adjustbox}
    \caption{Hyperparameters of the design choices discussed in this paper}
    \label{tab:hyperparameters}
\end{table}
\section{Further Experiments}

We conduct further experiments trying out different hyperparameters in the models which conditioned on local ResNet features. The results can be seen in Figure \ref{table:appendix_table}. The models used are as follows:

\textbf{MLP}
uses the MLP point encoder and 4x4 feature map from ResNet. The diffusion policy uses an embedding dimension of 128. 

\textbf{MLP256}
is similar to MLP but the diffusion policy has an embedding dimension of 256. 

\textbf{SUGAR}
uses the point cloud encoder from the SUGAR pretrained model and 4x4 feature map from ResNet. The point cloud encoder is frozen. The diffusion policy uses an embedding dimension of 128.

\textbf{SUGAR-FT2}
is similar to SUGAR but the last two layers are finetuned while keeping the other layers frozen.

\textbf{SUGAR256-FT2}
is similar to SUGAR-FT2 but the diffusion policy uses an embedding dimension of 256.

\textbf{MLP8x8}
uses the MLP point encoder and 4x4 feature map from ResNet. The transformer used to get the condition vector from the ResNet features has an embedding dimension of 256. The diffusion policy uses an embedding dimension of 128. 

\textbf{MLP8x8-L512}
is similar to MLP8x8 but the transformer used to get the condition vector from the ResNet features has an embedding dimension of 512.

\section{Adaptive LayerNorm conditioning}
\label{sec:adaln}

A visualization of the adaptive layer norm is given in Figure \ref{fig:dit_block}. We use the point cloud and language as primary modality in this visualization. In a Diffusion Transformer (DiT) block visualized in Figure \ref{fig:dit_block}, the most significant difference to a vanilla transformer block is scaling and shifting operations conditioned on the image CLS token. The scaling factors $\alpha$, $\gamma$ and the shifting factor $\beta$ are applied to self-attention and feed-forward part of the DiT block. The expression $\text{AdaLN}(z_t^P, z_t^L | z_t^I)$ indicates that image embedding is used as condition and mapped to factors $\alpha$, $\gamma$ and $\beta$, while the point cloud and language embeddings go through the self-attention and feed-forward blocks with additional scaling and shifting operations by these factors.

\begin{table*}[t!]
\begin{center}
\begin{sc}
\resizebox{0.9\textwidth}{!}{%
\setlength{\aboverulesep}{0pt}
\setlength{\belowrulesep}{0pt}
\renewcommand{\arraystretch}{1.5}
\begin{tabular}{l|cccccccc}
\toprule
\textbf{Task} & MLP & MLP256 & SUGAR & SUGAR-FT2 & SUGAR256-FT2 & MLP8x8 & MLP8x8-L512 \\
\midrule
PnPCabToCounter
& $0.16$
& $0.10$
& $0.04$
& $0.08$
& $0.10$
& $0.10$
& $0.16$
\\

PnPCounterToCab
& $0.08$
& $0.08$
& $0.04$
& $0.02$
& $0.14$
& $0.22$
& $0.08$
\\

PnPCounterToMicrowave
& $0.22$
& $0.20$
& $0.04$
& $0.08$
& $0.10$
& $0.18$
& $0.26$
\\

PnPCounterToSink
& $0.08$
& $0.08$
& $0.00$
& $0.00$
& $0.08$
& $0.06$
& $0.06$
\\

PnPCounterToStove
& $0.02$
& $0.06$
& $0.00$
& $0.02$
& $0.04$
& $0.04$
& $0.06$
\\

PnPMicrowaveToCounter
& $0.04$
& $0.08$
& $0.02$
& $0.06$
& $0.12$
& $0.10$
& $0.08$
\\

PnPSinkToCounter
& $0.24$
& $0.26$
& $0.08$
& $0.08$
& $0.30$
& $0.20$
& $0.22$
\\
PnPStoveToCounter
& $0.26$
& $0.28$
& $0.02$
& $0.04$
& $0.26$
& $0.18$
& $0.20$
\\
\midrule
OpenSingleDoor
& $0.62$
& $0.58$
& $0.52$
& $0.44$
& $0.74$
& $0.64$
& $0.68$
\\

OpenDoubleDoor
& $0.88$
& $0.94$
& $0.74$
& $0.70$
& $0.92$
& $0.90$
& $0.94$
\\
CloseDoubleDoor
& $0.84$
& $0.82$
& $0.56$
& $0.76$
& $0.78$
& $0.70$
& $0.82$
\\
CloseSingleDoor
& $0.80$
& $0.84$
& $0.68$
& $0.84$
& $0.84$
& $0.86$
& $0.86$
\\
\midrule
OpenDrawer
& $0.66$
& $0.68$
& $0.76$
& $0.84$
& $0.72$
& $0.60$
& $0.62$
\\

CloseDrawer
& $0.90$
& $0.96$
& $0.96$
& $0.96$
& $0.94$
& $0.96$
& $0.90$
\\
\midrule
TurnOnStove
& $0.56$
& $0.46$
& $0.62$
& $0.54$
& $0.66$
& $0.48$
& $0.46$
\\

TurnOffStove
& $0.14$
& $0.16$
& $0.22$
& $0.14$
& $0.20$
& $0.12$
& $0.12$
\\
\midrule
TurnOnSinkFaucet
& $0.40$
& $0.60$
& $0.68$
& $0.58$
& $0.70$
& $0.68$
& $0.68$
\\

TurnOffSinkFaucet
& $0.50$
& $0.80$
& $0.68$
& $0.82$
& $0.78$
& $0.76$
& $0.82$
\\

TurnSinkSpout
& $0.50$
& $0.52$
& $0.58$
& $0.60$
& $0.52$
& $0.60$
& $0.54$
\\
\midrule
CoffeePressButton
& $0.92$
& $0.90$
& $0.84$
& $0.92$
& $0.90$
& $0.84$
& $0.86$
\\

TurnOnMicrowave
& $0.76$
& $0.26$
& $0.62$
& $0.68$
& $0.68$
& $0.60$
& $0.74$
\\

TurnOffMicrowave
& $0.92$
& $0.68$
& $0.90$
& $0.82$
& $0.96$
& $0.82$
& $0.86$
\\
\midrule
CoffeeServeMug
& $0.50$
& $0.56$
& $0.56$
& $0.60$
& $0.48$
& $0.56$
& $0.62$
\\

CoffeeSetupMug
& $0.18$
& $0.14$
& $0.14$
& $0.14$
& $0.16$
& $0.20$
& $0.22$
\\
\midrule
\textbf{Average Success Rate}
& $0.4658$
& $0.4600$
& $0.4292$
& $0.4483$
& $0.5050$
& $0.4750$
& $0.4942$
\\
\bottomrule
\end{tabular}
}
\end{sc}
\end{center}
    \caption{Further results for RoboCasa with 50 Human Demonstrations conditioning on local ResNet features \\
    }
\label{table:appendix_table}
\end{table*}

\end{document}